\documentclass[11pt,a4paper]{article}
\usepackage[hyperref]{acl2021}
\usepackage{times}
\usepackage{latexsym}

\usepackage{multicol}
\usepackage{microtype}
\usepackage{graphicx}
\usepackage{booktabs}
\usepackage{multirow}
\usepackage{amsthm,amsmath,amssymb}
\usepackage{mathrsfs}
\usepackage{url}
\usepackage[linesnumbered,ruled,vlined,algo2e]{algorithm2e}
\usepackage{arydshln}
\usepackage{xspace}
\usepackage{CJK}
\usepackage{relsize}
\usepackage{times}
\usepackage{latexsym}
\usepackage{bm}
\usepackage{amsmath}
\usepackage{url}
\usepackage{makecell}
\usepackage{subfigure}
\usepackage{graphicx} 
\usepackage{colortbl}

\DeclareMathOperator*{\argmax}{arg\,max}
\renewcommand\vec[1]{\overrightarrow{#1}}
\newcommand\cev[1]{\overleftarrow{#1}}
\newcommand\bid[1]{\overleftrightarrow{#1}}
\aclfinalcopy 

\title{The USYD-JD Speech Translation System for IWSLT2021}

\author{
Liang Ding\\
The University of Sydney\\
\normalsize \texttt{ldin3097@sydney.edu.au}\\
\And
Di Wu\thanks{Work was done when Di Wu was visiting at JD.}\\
Peking University\\
\normalsize \texttt{inbath@163.com}\\
\And
Dacheng Tao\\
JD Explore Academy, JD.com\\
\normalsize \texttt{dacheng.tao@gmail.com}}

\date{}

\begin{document}
\maketitle
\begin{abstract}
This paper describes the University of Sydney \& JD's joint submission of the IWSLT 2021 low resource speech translation task. 
We participated in the Swahili$\rightarrow$English direction and got the best scareBLEU (25.3) score among all the participants.
Our constrained system is based on a pipeline framework, i.e. ASR and NMT. We trained our models with the officially provided ASR and MT datasets. 
The ASR system is based on the open-sourced tool Kaldi and this work mainly explores how to make the most of the NMT models. 
To reduce the punctuation errors generated by ASR model, we employ our previous work SlotRefine to train a punctuation correction model.
To achieve better translation performance, we explored the most recent effective strategies, including back translation, knowledge distillation, multi-feature reranking and transductive finetuning. For model structure, we tried auto-regressive and non-autoregressive models, respectively. In addition, we proposed two novel pre-train approaches, i.e. \textit{de-noising training} and \textit{bidirectional training} to fully exploit the data.
Extensive experiments show that adding the above techniques consistently improves the BLEU scores, and the final submission system outperforms the baseline (Transformer ensemble model trained with the original parallel data) by approximately 10.8 BLEU score, achieving the SOTA performance.
\end{abstract}

\section{Introduction}
Recent years have seen a surge of interest in speech translation (ST,~\citealt{ney1999speech}) task, that translates the source-side speech to the target-side text directly. The ST task contains two major components, Automatic Speech Recognition (ASR,~\citealt{jelinek1997statistical}) and Machine Translation (MT,~\citealt{koehn2009statistical}).
In this year's IWSLT low-resource speech translation task, our USYD-JD translation team participated in the Swahili to English track. We break the speech translation task into ``ASR$\rightarrow$NMT'' pipeline, and mainly focus on the NMT component. 

For model frameworks, we tried autoregressive neural machine translation, including Transformer-\textsc{Base} and -\textsc{Big}~\cite{transformer}, and non-autoregressive translation models~\cite{gu2018non}. Also, we employ our previous work SlotRefine~\cite{wu2020slotrefine} to tackle the case and punctuation problems after ASR. To make the most of the parallel and monolingual data, we proposed two pretrain strategies, i.e. \textsc{Bidirectional Pretraining}~\S\ref{subsec:bit} and \textsc{Denoising Pretraining}~\S\ref{subsec:denoise}, and employed two data augmentation strategies, i.e. \textsc{Bidirectional Self-Training}~\S\ref{subsec:bist} and \textsc{Tagged Back Translation}~\S\ref{subsec:tbt}. Where the data used for tagged back translation are carefully selected with our proposed multi-feature in-domain selection approach in \S\ref{subsec:data}. For post finetune/ process, we employed \textsc{Transductive Fine-Tune}~\S\ref{subsec:transfinetune} and a simple postprocessing approach~\S\ref{subsec:post}.

This paper is structured as follows: Section~\ref{sec:app} describes the major approaches we used. We present the data descriptions in Section~\ref{sec:data}. The experiments settings and main results are shown in Section~\ref{sec:exp}. Finnaly, we conclude our work in Section~\ref{sec:con}.

\section{Approaches}
\label{sec:app}

\subsection{Autoregressive Translation}
\label{subsec:at}
Given a source sentence $\bf x$, an NMT model generates each target word ${\bf y}_t$ conditioned on previously generated ones ${\bf y}_{<t}$. Accordingly, the probability of generating $\bf y$ is computed as:
\begin{equation}
    p({\bf y}|{\bf x})
    =\prod_{t=1}^{T}p({\bf y}_t|{\bf x},{\bf y}_{<t}; \theta)
    \label{eq:standard}
\end{equation}
where $T$ is the length of the target sequence and the parameters $\theta$ are trained to maximize the likelihood of a set of training examples according to $\mathcal{L}(\theta) = \argmax_{\theta} \log p({\bf y}|{\bf x}; \theta)$. Typically, we choose Transformer~\cite{transformer} as its SOTA performance. The training examples can be formally defined as follows:
\begin{equation}
    \vec{\text{B}} = \{(\mathbf{x}_i, \mathbf{y}_i)\}^N_{i=1}
    \label{eq:B}
\end{equation}
where $N$ is the total number of sentence pairs in the training data. Note that in standard MT training, the $\bf{x}$ is feed to the encoder and $\textbf{y}_{<t}$ to the decoder to finish the conditional estimation for $\textbf{y}_t$, thus the utilization of $\vec{\text{B}}$ is directional, i.e. $\mathbf{x}_i$$\rightarrow$$\mathbf{y}_i$. In the preliminary experiments, we utilized autoregressive translation (AT) model for \textit{translation}, \textit{case correction} and \textit{punctuation generation} tasks as its powerful modelling ability and generation accuracy.

\subsection{Bidirectional Pretraining}
\label{subsec:bit}
\paragraph{Motivation}
The motivation is when human learn foreign languages with translation examples, e.g. $\mathbf{x}_i$ and $\mathbf{y}_i$. Both directions of this example, i.e. $\mathbf{x}_i$$\rightarrow$$\mathbf{y}_i$ and $\mathbf{y}_i$$\rightarrow$$\mathbf{x}_i$, may help human easily master the bilingual knowledge. 
Motivated by this, \newcite{levinboim-etal-2015-model,Liang2007AgreementBasedL} propose to modelling the invertibility between bilingual languages. \newcite{cohn2016incorporating} introduce extra bidirectional prior regularization to achieve symmetric training from the point view of training objective. \newcite{he2018layer,zheng2019mirror} enhance the coordination of bidirectional corpus with model level modifications. Different from the above methods, we model both directions of a given training example by a simple data manipulation strategy.
\paragraph{Our Implementation}
Many studies have shown that pretraining could transfer the knowledge and data distribution, hence improving the generalization~\cite{hendrycks2019using,mathis2021pretraining}. Here we want to transfer the bidirectional knowledge among the corpus. Specifically, we propose to first pretrain MT models on bidirectional corpus, which can be defined as follows:
\begin{equation}
    \bid{\text{B}} = \{(\mathbf{x}_i, \mathbf{y}_i) \cup (\mathbf{y}_i, \mathbf{x}_i)\}^N_{i=1}
    \label{eq:bi.B}
\end{equation}
such that the $\theta$ in Equation~\ref{eq:standard} can be updated by both directions, then the bidirectional pretraining (BiPT) objective can be formulated as:
\begin{align}
    {\mathcal{L}_\text{BiPT}}(\theta) = &\overbrace{\argmax_{\theta} \log p({\bf y}|{\bf x}; \theta)}^{\text{Forward}: \vec{\mathcal{L}_{\theta}}}\\
    &+ \underbrace{\argmax_{\theta} \log p({\bf x}|{\bf y}; \theta)}_{\text{Backward}: \cev{\mathcal{L}_{\theta}}}
    \label{eq:bitloss}
\end{align}
where the $\text{forward}~\vec{\mathcal{L}_{\theta}}$ and $\text{backward}~\cev{\mathcal{L}_{\theta}}$ are optimized iteratively. 
From data perspective, we achieve the bidirectional updating as follows: 1) swapping the source and target sentences of a parallel corpus, and 2) appending the swapped version to the original. Then the training data was doubled to make better and full use of the costly bilingual corpus.
The pretraining can acquire general knowledge from bidirectional data, which may help {\em better} and {\em faster} learning further tasks. 
Thus, we early stop bidirectional training at 1/3 of the total steps. To ensure the proper training direction, we further train the pretrained model on required direction $\vec{\text{B}}$ with the rest of 2/3 training steps. Considering the effectiveness of pretraining~\cite{mathis2021pretraining} and clean finetuning~\cite{wu2019exploiting}, we introduce a combined pipeline: $\bid{\text{B}}\rightarrow\vec{\text{B}}$ as out best training strategy.

\subsection{Denoising Pretraining}
\label{subsec:denoise}
\paragraph{Motivation}
The motivation is when human learn one language, one of the best practices for language acquisition is to correct the sentence errors, e.g. ${noised}(\mathbf{x}_i)$$\rightarrow$$\mathbf{x}_i$ and ${noised}(\mathbf{y}_i)$$\rightarrow$$\mathbf{y}_i$. Motivated by this, \newcite{lewis2020bart} propose several noise adding approaches and denoise them with end-to-end pretraining. \newcite{liu2020multilingual} introduce this idea to the multilingual scenarios. Different from above monolinugal denoising pretraining approaches, we proposed a simpler noise function and apply them to each side of the parallel data.
\paragraph{Our Implementation}
Here we want the model to understand the source- and target-side languages well.
For noise function ${noised}(\cdot)$, we apply the common noise-injection practice, i.e. removing, replacing, or nearby swapping one time for a random word with a uniform distribution in a sentence~\cite{edunov2018understanding,ding2020self}. Then the size of the original parallel data doubled as follows:
\begin{align}
    \text{S}_{\mathrm{src}} &= \{{noised}(\mathbf{x}_i), \mathbf{x}_i\}^N_{i=1}\\
    \text{S}_{\mathrm{tgt}} &= \{{noised}(\mathbf{y}_i), \mathbf{y}_i\}^N_{i=1}
    \label{eq:denoise}
\end{align}
where $\text{S}_{\mathrm{src}}$ and $\text{S}_{\mathrm{tgt}}$ can be combined to update the end-to-end model to achieve denoising pretraining. such that the $\theta$ in Equation~\ref{eq:standard} can be updated by denoising both the source and target data, then the denoisig pretraining (DPT) objective can be formulated as:
\begin{align}
    {\mathcal{L}_\text{DPT}}(\theta) = &\overbrace{\argmax_{\theta} \log p({\bf x}|{{noised}({\bf x})}; \theta)}^{\text{Source Denoising}: {\mathcal{L}_{\theta}^{S}}}\\
    &+ \underbrace{\argmax_{\theta} \log p({\bf y}|{{noised}({\bf y})}; \theta)}_{\text{Target Denoising}: {\mathcal{L}_{\theta}^{T}}}
    \label{eq:dptloss}
\end{align}
where the ${\text{Source Denoising}: {\mathcal{L}_{\theta}^{S}}}$ and ${\text{Target Denoising}: {\mathcal{L}_{\theta}^{T}}}$ are optimized iteratively. The pretraining can store knowledge of the source and target languages into the shared model parameters, which may help {\em better} and {\em faster} learning further tasks. 
Similar to bidirectional pretraining in \S\ref{subsec:bit}, we early stop denoising training at 1/3 of the total steps, and tune the model normally with the rest of 2/3 training steps. This process can be formally denoted as such pipeline: $\text{S}_{\mathrm{src}}+\text{S}_{\mathrm{tgt}}\rightarrow\vec{\text{B}}$.

Note that Bidirectional Pretraining (BiPT) and Denoising Pretraining (DPT) can be combined and further enhance the model performance (The effect of their complementary can be found in Table~\ref{tab:result}). In particular, the combination order of BiPT and DPT are empirically inspired by human learning behavior, where a good interpreter will first master at least one language (usually the mother tongue), and then learn other languages and achieve bilingual translation. Thus, the combined pretraining process follows $\text{DPT}\rightarrow\text{BiPT}$. In combined pretraining setting, we will train longer until the model converges completely.

\subsection{Nonautoregressive Translation}
\label{subsec:nat}
Different from autoregressive translation~\cite[AT]{rnnsearch,transformer} models that generate each target word conditioned on previously generated ones, non-autoregressive translation~\cite[NAT]{gu2018non} models break the autoregressive factorization and produce the target words in parallel.
Given a source sentence $\bf x$, the probability of generating its target sentence $\bf y$ with length $T$ is defined by NAT as:
\begin{equation}
    p({\bf y}|{\bf x})=p_L(T|{\bf x}; \theta) \prod_{t=1}^{T}p({\bf y}_t|{\bf x}; \theta)
\end{equation}
where $p_L(\cdot)$ is a separate conditional distribution to predict the length of target sequence. 
Typicallly, most NAT models are implemented upon the framewok of Transformer~\cite{transformer}.
In the preliminary experiments, we utilized NAT for \textit{translation}, \textit{case correction} and \textit{punctuation generation} tasks as NAT can well avoid the error accumulation and exposure bias problems during generation. Also, we employ several advanced structure~\cite{gu2019levenshtein,ding-etal-2020-context} (\textit{Levenshtein} with source local context modelling) and our proposed training strategies~\cite{Ding2020Progressive,Ding2020Rejuvenating,Ding2020UnderstandingAI} as default settings.

\subsection{Bidirectional Self-Training}
\label{subsec:bist}
Besides improving NMT at model level, many researchers turn to data perspective, including exploiting the parallel and monolingual data. The most representative approaches include: a) {Back Translation} (\textbf{BT},~\citealt{sennrich2016improving}) combines the synthetic data generated with target-side monolingual data and parallel data; b) {Knowledge Distillation} (\textbf{KD},~\citealt{kim-rush-2016-sequence}) trains the model with sequence-level distilled parallel data; c) {data diversification} (\textbf{DD},~\citealt{nguyen2019data}) diversifies the data by applying KD and BT on parallel data. Clearly, self-training is at the core of above approaches, that is, they generate the synthetic data either from source to target or reversely, with either monolingual or bilingual data.

To this end, we propose a bidirectional self-training approach for both parallel and monolingual data (including source and target, respectively). Specifically, the base teacher models are trained with original parallel data in the first iteration (Round 1 in Table~\ref{tab:selftraining}), and based on these forward- and backward-teachers, all available Swahili \& English sentences can be used to generate the corresponding synthetic English \& Swahili sentences. After balanced-sampling between synthetic and authentic data, the concatenated data can be used to train the second iteration teachers (Round 2 in Table~\ref{tab:selftraining}).

To reveal why our approach works, we show the results in Table~\ref{tab:bi-selftraining} from the point view of data complexity~\cite{zhou2019understanding}. Self-training reduces the data complexity, thus increasing the model deterministic and in turn enhancing the model performance.

\subsection{Data Selection Features for Back Translation}
\label{subsec:data}

Inspired by~\citet{ding-tao-2019-university}, where their cycle-translation strategy (generating high quality in-domain data) for back translation obtain substantial gains, we carefully design criteria for choosing monolingual in-domain corpus. First, we employ rule-based features, language model features. The feature types are described in Table~\ref{tab:data-features}. Our BERT language model used here is trained from scratch by the open-source tool\footnote{\url{https://github.com/huggingface/pytorch-pretrained-BERT}} with target side data. The Moore-Lewis in-domain scoring strategy~\cite{moore-lewis-2010-intelligent} is used where the language model scores are trained with Transformer~\cite{transformer}. We score all sentences in non-autoregressive fashion\footnote{\url{https://github.com/alphadl/EasyScore}} to utilize contextualized information.

According to our observations, by using above multiple data selection filters, issues like illegal characters, unfluent and domain unmatched sentences could be significantly reduced. The data statistics for back translation monolingual data can be found in Table~\ref{tab:tagbtmono}.

\begin{table}[t!]
    \begin{center}
    \scalebox{0.91}{
    \begin{tabular}{|c|l|}
    \hline 
    \small{} & \small{Features}\\
    \hline
    \multirowcell{4}{\small{LM Features}} & \small{BERT LM}~\cite{devlin2018bert}\\
    \cline{2-2}
     & \small{Transformer LM}~\cite{bei-etal-2018-empirical}\\
     \cline{2-2}
     ~ & \small{N-gram LM}~\cite{stolcke2002srilm} \\
     \hline
     \small{In-domain features} & \small{Moore-Lewis}~\cite{moore-lewis-2010-intelligent}\\
     \hline
     \small{Rule-based features} & \small{Illegal characters~\cite{bei-etal-2018-empirical}}\\
     \hline
     \multirowcell{1}{\small{Count Features}} & \small{Word count}\\
     \hline
    \end{tabular}}
    \end{center}
    \caption{\label{tab:data-features} Features for back translation data selection.}
\end{table}

\subsection{Tagged Back Translation}
\label{subsec:tbt}
Back-translation~\cite{sennrich2016improving,bojar-etal-2018-findings}, translating the large scale monolingual corpus to generate synthetic parallel data by Target-to-Source pretrained model, has been widely utilized to improve the translation quality. However, recent studies find that back translation increase the target-original test set performance rather than source-original ones from the perspective of translationese\footnote{Source-Original denotes the testing data originating in the source language, while target-original denotes the data translating from the target language.}~\cite{zhang2019effect,graham-etal-2020-statistical}. To eliminate such concerns, we leverage tagged back translation~\cite{caswell2019tagged} to improve the source-original testing performance. The implementation is straightforward, that is, adding a simple tag on the beginning of each source-side synthetic sentence. The detailed reason why this trick works can be found in \citealp{marie2020tagged}.

To ensure tagged back translation works well for our task, we carefully selected the target side in-domain monolingual data (\S\ref{subsec:data}). Final results in Table~\ref{tab:result} show the effectiveness of tagged back translation $\#9$ against competitive model $\#8$ (+1.9 BLEU scores).

\begin{table}[t!]
    \begin{center}
    \begin{tabular}{l|p{5.74cm}}
    \toprule
    src & \textit{Msimu uliopita wa Siltala kwenye ligi ilikuwa \bf{2006-07}}\\ 
    \midrule
    pred & \textit{Siltala's previous season in the league was \bf{2006 at 07}}\\ 
    \midrule
    +post & \textit{Siltala's previous season in the league was \bf{2006-07}}\\
    \bottomrule
    \end{tabular}
    \end{center}
    \caption{\label{tab:post-process}Example of the effectiveness of post-processing in handling inconsistent number translation.}
\end{table}

\subsection{Transductive Fine-Tuning}
\label{subsec:transfinetune}
The key idea of transductive finetune is that source input sentences from the validation and test sets are firstly translated to the target language space with the best well-performed NMT model, which results in a pretranslated synthetic dataset. Then models are finetuned on the generated synthetic dataset. We borrow this concept from previous systems~\cite{wu2020tencent,wang-etal-2020-tencent}. We empirically show that transductive finetune ($\#10-11$ in Table~\ref{tab:result}) indeed improves the official validation performance but harms the performance of our sampled valid\& test set that co-distributed with the training set. Note that we randomly sampled 5K/ 5K sentences from the training set as valid and test sets, respectively, to avoid the sub-optimal problem caused by the distribution gap. Experimental details can be found in \S\ref{sec:data} and \ref{sec:exp}.

\subsection{Reranking N-best Hypotheses}
\label{subsec:rerank}
As the NMT decoding being generally from left to right, this leads to label bias problem~\cite{lafferty2001conditional}. To alleviate this problem, besides using NAT (\S\ref{subsec:nat}),
we rerank the n-best hypotheses through training a $k$-best batch MIRA ranker~\cite{Cherry:2012:BTS:2382029.2382089} with multiple features on validation set. The feature pool we integrated include R2L (right-to-left) translation model, T2S (target-to-source) translation model, language model and IBM model 2 alignment score. After multi-feature reranking, the best hypothesis was retained.

\paragraph{Right-to-Left NMT Model}
The R2L NMT model using the same training data but with inverted target sentences (\textit{i.e.}, reverse target side characters ``a b c d''$\rightarrow$``d c b a''). Then, inverting the hypothesis in the $n$-best list can obtain perplexity score by R2L model.

\paragraph{Target-to-Source NMT Model}
The T2S model was initially trained for back-translation, we can employ this model to assess the translation adequacy as well by adding the T2S feature to reranking feature pool.

\paragraph{Language Model}
Besides above features, we employ language models as an auxiliary feature to give the fluent sentences better scores such that the results are easier to understand by human.

\subsection{Post Processing}
\label{subsec:post}
Besides general post-processing (\textit{i.e.}, de-BPE, de-tokenization and de-truecase~\footnote{\url{https://github.com/moses-smt/mosesdecoder/tree/master/scripts}}), we also used a post-processing algorithm~\cite{wang-etal-2018-niutrans} for inconsistent number, date translation, for example, ``\textit{2006-07}'' might be segmented as ``\textit{2006 -@@ 07}'' by BPE, resulting in the wrong translation ``\textit{2006 at 07}''. Our post-processing algorithm will search for the best matching number string from the source sentence to replace these types of errors, see Table~\ref{tab:post-process}.

\section{Data Preparation}
\label{sec:data}
\begin{figure*}
    \centering
    \includegraphics[width=0.95\textwidth]{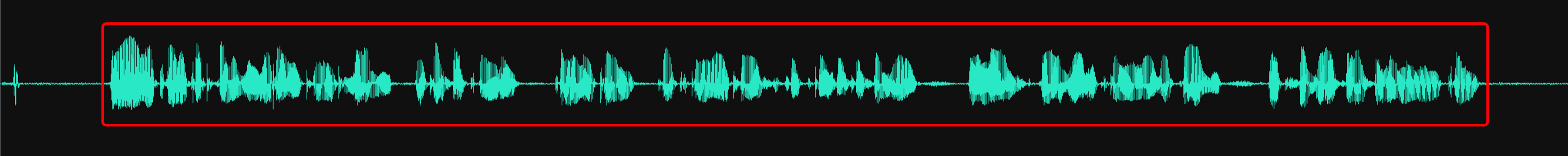}
    \caption{An example of how our rule-based voice activity detection model works on the waveform. Note that only the part in the red box will be retained as a valid fragment.}
    \label{fig:sound}
\end{figure*}

For ASR task, we downloaded all available Swahili speech-to-text data\footnote{\url{https://iwslt.org/2021/low-resource}}, such as openslr\footnote{\url{https://www.openslr.org/25/}} and IARPA Babel\footnote{\url{https://catalog.ldc.upenn.edu/LDC2017S05}} etc., as training corpus and employ all default settings in Kaldi\footnote{\url{https://github.com/kaldi-asr/kaldi}} to preprocess and train them. To simplify the ASR task, we lowercased all Swahili sentences and removed punctuation. To rejuvenate these case and punctuation information, we design two pipeline tasks after ASR: \textit{case correction} task and \textit{punctuation generation}. Also, it is worth noting that we design some rules to perform the ``voice activity detection'' process for the official speech testset. Take a piece of speech in Figure~\ref{fig:sound} for example, partial of speech in the red box will be keep as the valid input.

For NMT task, the parallel datasets we utilized are described at Table~\ref{tab:parall}, including CCAligned~\cite{elkishky_ccaligned_2020}, Tanzil~\cite{TIEDEMANN12.463}, ParaCrawl~\footnote{\url{https://www.paracrawl.eu/index.php}}, WikiMatrix~\cite{schwenk2019wikimatrix}, GlobalVoices~\cite{TIEDEMANN12.463}, TED2020~\cite{reimers-2020-multilingual-sentence-bert}, WikiMedia~\cite{TIEDEMANN12.463} and Gamayun~\footnote{\url{https://gamayun.translatorswb.org/data/}}. The monolingual data we utilized are described in Table~\ref{tab:mono} and Table~\ref{tab:tagbtmono}, where the monolingual data in Table~\ref{tab:mono} are used to train the system $\#1-8$ in Table~\ref{tab:result}, and data in Table~\ref{tab:tagbtmono} are used to train the system $\#9-11$ in Table~\ref{tab:result}, respectively. Table~\ref{tab:selftraining} denotes how the data used and generated by iterative bidirectional self-training (\S\ref{subsec:bist}). The total data size after two round of bidirectional self-training is 50.4M, and after tagged back translation, the final data volume is 60.4M.

To avoid the sub-optimal problem caused by the distribution gap between official validation and training data, we randomly sampled 5K/ 5K sentences from the training set as valid and test sets, respectively. The randomly sampled valid sentences are used to optimize the hype-parameters.

\begin{table}
    \begin{center}
    \begin{tabular}{|l|r|}
    \hline
    \textbf{Available Parallel Corpus} & \textbf{\#Sent.}\\
    \hline
    CCAligned& 2,044,993 \\
    Tanzil& 138,253 \\
    ParaCrawl& 132,517 \\
    WikiMatrix& 51,387 \\
    GlobalVoices& 32,307 \\
    TED2020& 9,754 \\
    Gamayun& 5,000 \\
    WikiMedia& 771 \\
    \hline
    Total & 2,414,982\\
    \hline
    \end{tabular}
    \end{center}
    \caption{\label{tab:parall}Statistics of parallel data.}
\end{table}

\begin{table}
    \begin{center}
    \begin{tabular}{|l|r|}
    \hline
    \textbf{Sampled Mono. Corpus} & \textbf{\#Sent.}\\
    \hline
    commoncrawl English& 4,366,344 \\
    commoncrawl Swahili& 38,928 \\
    ~~~+ upsampling (14$\times$)& 544,992\\
    \hline
    \end{tabular}
    \end{center}
    \caption{\label{tab:mono}Statistics of monolingual data.}
\end{table}

\begin{table}
    \begin{center}
    \begin{tabular}{|l|r|}
    \hline
    \textbf{Mono. Corpus for Tagged BT} & \textbf{\#Sent.}\\
    \hline
    \multicolumn{2}{|c|}{Totally collected corpus}\\
    \hline
    commoncrawl English& 30,513,498 \\
    \hline
    \multicolumn{2}{|c|}{Cleaned corpus with criteria in \S\ref{subsec:data}}\\
    \hline
    in domain English& 10,000,000 \\
    \hline
    \end{tabular}
    \end{center}
    \caption{\label{tab:tagbtmono}Statistics of monolingual data for Tagged Back-Translation.}
\end{table}

\section{Experiments}
\label{sec:exp}
\paragraph{Settings}
For \textit{case correction} and \textit{punctuation generation} tasks mentioned in \S\ref{sec:data}, we tried Autoregressive Transformer-\textsc{Base}~(AT, \S\ref{subsec:at}), Non-Autoregressive model~(NAT, \S\ref{subsec:nat}) and our previously designed \textsc{SlotRefine}~\cite{wu2020slotrefine}. In our preliminary experiments, NAT and SlotRefine work better on \textit{case correction} and \textit{punctuation generation} tasks, respectively, thus leaving as the default components in our final speech translation pipeline.

\begin{table}
    \begin{center}
    \scalebox{0.91}{
    \begin{tabular}{|l|l|r|}
    \hline
    \bf \# & \textbf{Data Statistics} & \textbf{\#Sent.}\\
    \hline
    \multicolumn{3}{|c|}{Preparing for Self-Training}\\
    \hline
    1 & parallel English& 2.4M\\
    2 & parallel Swahili& 2.4M\\
    3 & monolingual English & 4.4M\\
    4 & monolingual Swahili & 0.4M\\
    \hline
    \multicolumn{3}{|c|}{Self-Training Round 1}\\
    \hline
    5 & synthetic parallel & 9.6M\\
    6 & authentic parallel & 2.4M\\
    7 & ~~~+ upsampling (4$\times$) & 9.6M\\
    \hline
    8 & concat \#5 and \#7& 19.2M\\
    \hline
    \multicolumn{3}{|c|}{Self-Training Round 2}\\
    \hline
    9 & refined parallel \#8 & 19.2M \\
    \hline
    10 & concat \#8 and \#9 &38.4M\\
    11 & upsampled authentic parallel \#6 ($5\times$) &12.0M\\
    \hline
    12 & concat \#10 and \#11 & 50.4M\\
    \hline
    \end{tabular}}
    \end{center}
    \caption{\label{tab:selftraining}Data statistics for bidirectional self-training. Note that \#5 ``synthetic parall'' comes from monolingual English (\#3 ), monolingual Swahili (\#4), parallel English (\#1), and parallel Swahili (\#2). In our preliminary experiments, 4$\times$ (\#7) and 5$\times$ (\#11) upsampling strategies perform best in their corresponding settings, thus leaving as the default settings.}
\end{table}
For NMT task, we tried Autoregressive Transformer-\textsc{Big}~(AT, \S\ref{subsec:at}) and Non-Autoregressive model~(NAT, \S\ref{subsec:nat}) in preliminary experiments, and found that AT performs robust on all settings. Thus we employ Transformer-\textsc{Big} for all MT systems. Inspired by \citet{he2019control}, we empirically adopt large batch strategy~\cite{edunov2018understanding} (i.e. 458K tokens/batch) to optimize the performance. The learning rate warms up to $1\times10^{-7}$ for 10K steps, and then decays for 30K (data volumes range from 2M to 10M) / 50K (data volumes large than 10M) steps with the cosine schedule. For regularization, we tune the dropout rate from [0.1, 0.2, 0.3] based on validation performance, and apply weight decay with 0.01 and label smoothing with $\epsilon$ = 0.1. We use Adam optimizer ~\citep{kingma2015adam} to train models. We evaluate the performance on an ensemble of last 10 checkpoints to avoid stochasticity.

For fair comparison, the metric we employed is sacreBLEU~\cite{post-2018-call}. Training set, validation set and test set are processed consistently. Both Swahili and English sentences are performed tokenization and truecasing with Moses scripts~\cite{Koehn2007MosesOS}. In order to limit the size of vocabulary of NMT models, we adopted byte pair encoding (BPE)~\cite{sennrich2016improving} with 32k operations. Larger beam size may worsen translation quality~\cite{koehn-knowles-2017-six}, thus we set beam\_size=10 when performing n-best reranking (\S\ref{subsec:rerank}). All models were trained on 4 16GB \verb|NVIDIA V100| GPUs.

\begin{table*}[hbt]
    \begin{center}
    \begin{tabular}{l|l|ccc||cl}
    \hline
    \hline
    \textbf{\#} & \textbf{Models} & Valid & Test & $\Delta_{ave}$ & Off. Valid & $\Delta$\\ 
    \hline
    $1$ & \textbf{Baseline} (w/ Para. Data) & 47.1 & 48.5 & $-$ & 31.8 & $-$\\
    $2$ & ~~~\verb|+Bidrectional Pretrain| & 48.5 & 49.9 & &\\
    $3$ & ~~~\verb|+Denoising Pretrain| & 48.6 & 49.6 &  &\\
    $4$ & ~~~\verb|+Combination of #2 and #3| & 48.9 & 50.1 & $+1.7$ &\\
    \hline
    $5$ & \textbf{Bi. Self-Training} (w/ Mono. \& Para. Data) & 49.4 & 50.8 & $+2.3$ &\\
    $6$ & ~~~\verb|+Combination of #2 and #3| & 50.1 & 51.6 & $+3.1$ &\\
    \hline
    $7$ & \bf Iterative Bi. Self-Training & 49.7 & 50.9 & $+2.5$ &\\
    $8$ & ~~~\verb|+Combination of #2 and #3| & 50.5 & 51.8 & $+3.4$ &38.2 & $+6.4$\\
    \hline
    $9$ & \bf $\text{\#8}$ + Tagged Back Translation & 52.4 & 53.1 & $+5.0$ &40.1 & $+8.3$\\
    \hline
    $10$ & \bf $\text{\#9}$ + Transductive Finetune & 51.8 & 53.0 & $+4.6$ &41.5 & $+9.7$\\
    $11$ & ~~~+Iterative \verb|+#10| & 51.6 & 52.8 & $+4.4$ &41.9 & $+10.1$\\
    \hline
    $12$ & \bf $\text{\#11}$ + Reranking & 52.1 & 53.5 & $+5.0$ & 42.3 & $+10.5$\\
    $13$ & \bf $\text{\#12}$ + Post Processing & 52.5 & 54.0 & $+5.5$ & 42.6 & $+10.8$\\
    \hline
    \multicolumn{7}{c}{SacreBLEU of Final Submission ($\text{\#13}$) on official test set \textbf{25.3}}\\
    \hline
    \hline
    \end{tabular}
    \end{center}
    \caption{\label{tab:result}{Sacrebleu} of Sw$\rightarrow$En on our randomly sampled ``Valid/ Test'' sets and official validation set ``Off. Valid'', where ``$\Delta$'' represents the performance gains compared with baseline $\text{\#1}$. The submitted system is $\text{\#13}$.}
\end{table*}

\begin{table}[t]
\centering
    \begin{tabular}{l|cc}
    \hline\hline
    \textbf{Data} & \textbf{Compl.} & \textbf{BLEU}\\
    \hline
    Baseline&7.87&47.1\\
    Bi. Self-Training&5.34&49.4\\
    Iterative Bi. Self-Training&\textbf{4.89}&\textbf{49.7}\\
    \hline
    \hline
    \end{tabular}
    \caption{Explanation of why Bidirectional Self-Training works. The data complexity ``Compl.'' is measured on their corresponding training sets and alignment information is trained with \textit{fast-align}~\cite{dyer-etal-2013-simple}. The BLEU scores are reported on our sampled validation set. }
    \label{tab:bi-selftraining}
\end{table}

\paragraph{Main results}
Our main experiment is shown in Table~\ref{tab:result}, our baseline system is developed with the original parallel corpus and last-$10$ ensemble strategy. Unsurprisingly, the baseline system relatively performs the worst. 

The proposed \textit{Bidirectional Pretrain} in \S\ref{subsec:bit} and \textit{Denoising Pretrain} in \S\ref{subsec:denoise} could consistently and significanly improve the model performance, showing their effectiveness in low resource scenarios~\cite{zhang2020empowering}. 
Clearly, combining \textit{Bidirectional Pretrain} and \textit{Denoising Pretrain} could achieve better results (averaged +1.7 BLEU scores), indicating their complementary. 

As shown in $\#5$ and $\#7$, the proposed \textit{Bidirectional Self-Training} and its refined iterative version, could consistently enhance the model. To explore why self-training improves model performance, we discuss it from the point view of data complexity. As shown in Table~\ref{tab:bi-selftraining}, with the \textit{Bidirectional Self-Training} iteratively progresses, the data complexity becomes lower, leading to the better BLEU scores.
Notably, the combination of our proposed two pretraining approaches push the SOTA performance up to higher points. We believe that the effect of our proposed two pretrain strategies are still under-investigated, which will leave as future works. 
Overall, with strategies $\#2-8$, the model performance in terms of official validation test achieves surprisingly \textbf{+6.4} BLEU scores.

The \textit{Tagged Back Translation} (\S\ref{subsec:tbt}) with in-domain monolingual data significantly improves the performance of both our sampled test set and official valid set by +5.0 and +8.3 against baseline, respectively.

We empirically show that \textit{Transductive FineTune} (\S\ref{subsec:transfinetune}) indeed improves the official validation performance but harms the performance of our sampled valid\& test set that co-distributed with the training set. This indicates that tranductive learning is a effective practice to transfer a well-trained model across domains.

And the last two strategies \textit{Reranking} (\S\ref{subsec:rerank}) and \textit{Post Processing} (\S\ref{subsec:post}) could further improve the official validataion BLEU score from 41.9 to 42.6, which substantially outperforms the baseline by +10.8 BLEU score.

\section{Conclusion and Future Work}
\label{sec:con}
This paper presents the University of Sydney \& JD's speech machine translation system for IWSLT2021  Swahili$\rightarrow$English task. The whole system is pipelined, containing ASR, case correction, punctuation generation and NMT tasks, and we main focused on NMT task.

We leveraged multi-dimensional strategies and frameworks to improve the translation qualities, which achieves surprisingly \textbf{+10.8 BLEU} scores improvement against baseline and ranks the \textbf{1st} among all the participants.
We find that our proposed \textsc{bidirectional pretraining} (\S\ref{subsec:bit}) and \textsc{denoising pretraining} (\S\ref{subsec:denoise}) can consistently improves the competitive baselines. Also, we employ \textsc{Bidirectional Self-Training} in \S\ref{subsec:bist} and \textsc{Tagged BT} in \S\ref{subsec:tbt} make the most of the existing parallel and monolingual data.

In the future, we would like to polish other components in the pipeline to achieve better performance. Also, it is worthy to try an end-to-end approach with cross-modal structures to incorporate audio and vision knowledge~\cite{xu2021vitae}. For robust model training and data utilization, we would explore better strategies, e.g. adversarial training~\cite{wu2021bridging} and curriculum learning~\cite{liu2020norm,zhou2021self}.

\section*{Acknowledgments}
We would thank anonymous reviewers for their considerate proofreading and valuable comments.

\bibliographystyle{acl_natbib}
\bibliography{acl2021}




\end{document}